\documentclass[letterpaper, 10 pt, conference]{ieeeconf}  % Comment this line out if you need a4paper

\IEEEoverridecommandlockouts                              % This command is only needed if 
\usepackage{times}
\usepackage{epsfig}
\usepackage{graphicx}
\usepackage{multirow}
\usepackage{amsmath}
\usepackage{amssymb}
\usepackage[nolist,nohyperlinks]{acronym}
\usepackage{todonotes}
\usepackage{subfig} 
\usepackage{alphalph} 
\usepackage{microtype}
\mathchardef\mhyphen="2D

\def\etal{\emph{et al}.\ }

\DeclareRobustCommand{\mysmallmath}[1]{\scriptscriptstyle{#1}}

\newcommand{\brackets}[3]{\ensuremath{\left#1 #2 \right#3}}
\newcommand{\mypar}[1]{\brackets{(}{#1}{)}}

\newcommand{\idx}[2][]{\ifthenelse{\equal{#1}{}}{#2}{#2{#1}}}

\newcommand{\mysym}[5]{\ensuremath{\prescript{#1}{#2\hspace{+0.25mm}}{#3}_{#4}^{#5}}}%Everything should be based on this

\newcommand{\myvtr}[1]{\ensuremath{\mathbf{\lowercase{#1}}}}

\newcommand{\vtr}[5]{\mysym{#1}{#2}{\myvtr{#3}}{#4}{#5}}

\newcommand{\stt}[0]{p}
\newcommand{\state}[4]{\vtr{}{}{\stt}{}{}}

\newcommand{\mask}[2]{M_{#2}^{#1}}

\newcommand{\imgset}[1]{\mathcal{Z}_{#1}}
\newcommand{\imgsetwto}[0]{\imgset{c}}
\newcommand{\imgsetwth}[0]{\imgset{c'}}

\newcommand{\imgcls}[1]{z_{#1}}
\newcommand{\imgwto}[0]{\imgcls{c}}
\newcommand{\imgwth}[0]{\imgcls{c'}}

\newcommand{\imgnovr}[0]{x_{c}}
\newcommand{\imgovr}[0]{x_{c'}}

\newcommand{\oscarcentered}[1]{\begin{tabular}{c} #1 \end{tabular}}

\title{Improving Robot Localisation by Ignoring Visual Distraction}

\author{Oscar Mendez \hspace{1cm} Matthew Vowels \hspace{1cm} Richard Bowden \thanks{All authors with the University of Surrey \tt\small \{o.mendez, m.vowels, r.bowden\}@surrey.ac.uk} \thanks{We acknowledge the support of the InnovateUK Autonomous Valet Parking Project Grant No 104273 and the EPSRC Project ROSSINI (EP/S016317/1).}
}

\begin{document}

\begin{acronym}[RANSAC] %replace parameter with longest acronym in list
%Mine 
\acro{nbv}[NBV]{Next-Best View}
\acro{nbs}[NBS]{Next-Best Stereo}

%Conferences
\acro{bmvc}[BMVC]{British Machine Vision Conference}
\acro{iccv}[ICCV]{International Conference on Computer Vision}

%Features
\acro{agast}[AGAST]{Adaptive and Generic Accelerated Segment Test}
\acro{fast}[FAST]{Features from Accelerated Segment Test}

%Metrics
\acro{ate}[ATE]{Absolute Trajectory Error}
\acro{rpe}[RPE]{Relative Pose Error}
\acro{rmse}[RMSE]{Root Mean Square Error}
\acro{mse}[MSE]{Mean Square Error}
\acro{psnr}[PSNR]{Peak Signal-to-Noise Ratio}

%Spaces
\acro{dof}[DoF]{Degrees of Freedom}
\acro{se2}[SE(2)]{Special Euclidean Space}
\acro{so2}[SO(2)]{Special Orthogonal Space}
\acro{se3}[SE(3)]{Special Euclidean Space}
\acro{so3}[SO(3)]{Special Orthogonal Space}

%Concepts
\acro{cnn}[CNN]{Convolutional Neural Network}
\acro{gan}[GAN]{Convolutional Neural Network}
\acro{lut}[LUT]{Lookup Table}
\acro{ls}[LS]{Least-Squares}
\acro{kdt}[k-d tree]{k-dimensional Tree}
\acro{if}[IF]{Information Filter}
\acro{lls}[L-LS]{Linear Least-Squares}
\acro{ills}[ILLS]{Iterative Linear-Least-Squares}
\acro{mvs}[MVS]{Multi-View Stereo}
\acro{mocap}[MoCap]{Motion Capture}
\acro{pid}[PID]{Proportional Integral Controller}
\acro{vp}[VP]{Vanishing Point}
\acro{svd}[SVD]{Single Value Decomposition}
\acro{ba}[BA]{Bundle Adjustment}
\acro{sf}[SF]{Sensor-Fusion}
\acro{sfm}[SfM]{Structure from Motion}
\acro{csfm}[CSfM]{Collaborative Structure from Motion}
\acro{vo}[VO]{Visual Odometry}

%Kalman
\acro{klf}[KF]{Kalman Filter}
\acro{ekf}[EKF]{Extended Kalman Filter}
\acro{ukf}[UKF]{Unscented Kalman Filter}

%Algorithms
  \acro{icp}[ICP]{Iterative Closest Point}
  \acro{lsd}[LSD]{Line Segment Detector}
  \acro{pnp}[PnP]{Perspective N-Points}
%  \acro{ransac}[RANSAC]{RANdom Sample And Consensus}
  \acro{rsc}[RANSAC]{RANdom SAmple and Consensus}
  \acro{svo}[SVO]{Semi-Direct Visual Odometry}
  \acro{vsfm}[VSFM]{Visual Structure from Motion}
  \acro{mse}[MSE]{Mean Squared Error}
  
  %SLAM
  \acro{slam}[SLAM]{Simultaneous Localisation and Mapping}
  \acro{vslam}[VSLAM]{Vision-Based Simultaneous Localisation and Mapping}
  
  %TAM
  \acro{etam}[ETAM]{ETHZASL-PTAM}
  \acro{ptam}[PTAM]{Parallel Tracking and Mapping}
  \acro{dtam}[DTAM]{Dense Tracking and Mapping}

  %MCL
  \acro{pf}[PF]{Particle Filter}
  \acro{rbpf}[RBPF]{Rao-Blackwellised Particle Filter}
  \acro{smc}[SMC]{Sequential Monte-Carlo}
  \acro{mkl}[MKL]{Markov Localisation}
  \acro{gbl}[GBL]{Grid-Based Localisation}
  \acro{mcl}[MCL]{Monte-Carlo Localisation}
  \acro{amcl}[AMCL]{Adaptive Monte Carlo Localisation}
  \acro{vmcl}[VMCL]{Vision-Based Monte-Carlo Localisation}
  \acro{rmcl}[RMCL]{Range-Based Monte-Carlo Localisation}
  
  %Pathplanners
  \acro{prm}[PRM]{Probabilistic Road Map}
  \acro{rrt}[RRT]{Rapidly-exploring Random Tree}
  \acro{rrt*}[RRT*]{Rapidly-exploring Random Tree}
   
  %Networks
  \acro{vae}[VAE]{Variational Autoencoder}
  \acro{vqvae}[VQ-VAE]{Vector Quantised \acs{vae}}
 
%Libraries
\acro{ompl}[OMPL]{Open Motion Planning Libraries}
\acro{pcl}[PCL]{Point Cloud Library}
\acro{ros}[ROS]{Robot Operating System}

%Sensors
\acro{gps}[GPS]{Global Positioning System}
\acro{imu}[IMU]{Inertial Measurement Unit}
\acro{lidar}[LiDAR]{Light Detection And Ranging}
\acro{rgb}[RGB]{Red, Green and Blue}
\acro{rgbd}[RGB-D]{\acs{rgb} and Depth}
\acro{rgbds}[RGB-DS]{\acs{rgb}, Depth and Semantic}
\acro{sedar}[SeDAR]{Semantic Detection and Ranging}
\acro{smcl}[SeMCL]{Semantic Monte-Carlo Localisation}
\acro{sonar}[SoNAR]{Sound Navigation And Ranging}

%Misc
\acro{ada}[ADA]{ARDrone Autonomy}
\acro{ai}[AI]{Artificial Intelligence}
\acro{ar}[AR]{Augmented Reality}
\acro{ard}[ARD]{AR Drone}
\acro{mct}[MTC]{MATLAB Calibration Toolbox}
\acro{tma}[TMA]{TUM\_ARDrone}
\acro{tum}[TUM]{Technical University of Munich}
\acro{uav}[UAV]{Unmanned Aerial Vehicle}
\acro{sdk}[SDK]{Software Development Kit}
\acro{fps}[FPS]{Frames Per Second}
\end{acronym}

\maketitle
\thispagestyle{empty}
\pagestyle{empty}

%%%%%%%%%%%%%%%%%%%%%%%%%%%%%%%%%%%%%%%%%%%%%%%%%%%%%%%%%%%%%%%%%%%%%%%%%%%%%%%%
\begin{abstract}
 Attention is an important component of modern deep learning.
 However, less emphasis has been put on its inverse: ignoring distraction.
 Our daily lives require us to explicitly avoid giving attention to salient visual features that confound the task we are trying to accomplish.
 This visual prioritisation allows us to concentrate on important tasks while ignoring visual distractors.
 
 In this work, we introduce Neural Blindness, which gives an agent the ability to completely ignore objects or classes that are deemed distractors.
 More explicitly, we aim to render a neural network completely incapable of representing specific chosen classes in its latent space.
 In a very real sense, this makes the network ``blind'' to certain classes, allowing and agent to focus on what is important for a given task, and demonstrates how this can be used to improve localisation.
 
\end{abstract}

\section{Introduction}

The presence of extraneous, confusing or overwhelming information in an image has been shown to reduce task-specific performance in humans \cite{mcmains2011}.
As such, it should not come as a surprise that similar performance drops are observed in machine learning applications.
From localisation in crowded scenes to tracking in environments with multiple objects, visual clutter can have a significant effect on the performance of neural networks.
This can be explained, in both humans and \ac{ai}, as multiple objects competing for neural representation \cite{mcmains2011}.

Object removal is an application in its own right, where the task is to fill in a ``gap'' left when an object is removed from an image. 
That removal is normally defined by a mask, which is required at runtime.
Object removal can be used to remove visual clutter from images as a pre-processing step to any task.
However, there are important limitations to this approach.
Firstly, it requires expensive image processing to be done for every single datum in a dataset and, crucially, also at run-time.
Secondly, it requires manual mask annotation of the objects that are to be removed.
Finally, the approach must reliably fill the image with temporally consistent content in order to be useful for down stream tasks.
Our key insight is that for some applications, it is less important that an object is removed from an image and more important that the computer is capable of completely ignoring that object.
In this work, we propose an approach to allow a neural network to ignore chosen objects, not by simply removing them from the image, but by training the network to not ``see'' them in the first place.

Our approach, which we call \textit{Neural Blindness}, aims to prevent a neural network from representing chosen objects in its latent space.
Humans are capable of ignoring clutter when the task at hand requires them to do so.
In a similar way, we train an autoencoder network to remove class-specific objects from its latent representation and demonstrate how this improves the performance of an autonomous agent.

We introduce two key novelties.
In the first, we introduce the concept of Neural Blindness along with a novel Siamese autoencoder architecture used for training neural-blind latent representations.
We additionally offer insights into why certain types of autoencoders are suitable for this task, while others are not.
In the second, we demonstrate our neural-blind latent representation can be used to improve the performance on downstream tasks by ignoring visual clutter.
More explicitly, we demonstrate Neural Blindness can be used as an effective pre-training approach to inject invariance to distractors, particularly when applied to localisation of autonomous vehicles.

\section{Literature Review} \label{sec:lit}
In many computer vision tasks involving the use of latent spaces has become commonplace \cite{ham2018, xiong2019, zadeh2019}. 
These latent spaces encode high-level semantic information about the image.
Our task aims for the removal of targeted information/objects from the original image such that this information is never encoded in the representation to begin with. 
The task therefore requires learning invariant representations and the ``forgetting'' of information. 

One way to obtain a latent space that is domain-specific or, equivalently, invariant to certain information, is via the use of \acp{vae} \cite{JADE, moyer1, DIVA, kingma}.  
A  \ac{vae} \cite{kingma} is a Bayesian autoencoder, whereby the training objective involves reconstruction of a high-dimensional observational distribution from a low-dimensional representation. 
By regularizing this representation so that it resembles a pre-specified prior distribution (often an isotropic Gaussian), the network undertakes variational inference and learns a smooth, traversible, latent manifold.

\acp{vae} are extremely flexible and have been applied to a broad range of tasks including the generation of high quality images \cite{razavi2019generating}, disentangled representation learning \cite{jiang2019, locatello2019}, object detection \cite{li2019}, motion synthesis \cite{habibie2017}, and  neural image inpainting \cite{ham2018, zadeh2019}.

Researchers have encouraged invariance or forgetting with \acp{vae} by utilizing various combinations of weak supervision \cite{JADE, moyer1}, adversarial training \cite{ganin1}, and by trading off reconstruction error against encoding capacity according to the information bottleneck principle \cite{tishby2015}. 
Others have sought to fully disentangle the latent space into independent partitions \cite{kulkarni}, thereby enabling the isolation of wanted from unwanted information. 
Whilst unsupervised methods for disentanglement exist \cite{disentanglement, dupont2018}, recent research indicates that such methods do not achieve disentanglement consistently \cite{locatello} and, even if they do, they do not allocate factors to the latent space in a predictable way. 
Furthermore, in order for \acp{vae} to handle complex data and to reconstruct detailed images, the complexity of the latent space is increased, reducing the emphasis on achieving disentangled and semantically interpretable latent spaces \cite{BIVA, Vahdat2020NVAE}. 
For instance, \ac{vqvae} \cite{razavi2019generating, Vandenoord2018vqvae} deviates from the typical formulation of a \ac{vae} and utilizes a deterministic, discrete, vector-quantized codebook as a latent space.   
In this work, we incorporate a number of techniques into the \ac{vae}/autoencoder learning in order to achieve a representation that is invariant to targeted objects.
 
The latent representations in neural networks can be used for downstream tasks such as classification and regression.
In this area, one of the key applications is camera pose regression and the most common family of approaches is PoseNet \cite{Kendall2015Posenet,kendall2016modelling, kendall2017geometric, brahmbhatt2018geometry} and its derivatives.
Fundamentally, these approaches rely on sensor/pose pairs to train a \acs{cnn} that can regress pose given an image. 
Normally, they start with a pre-trained encoder network and modify the final fully connected layers to regress 6-\ac{dof} camera poses.
The network then learns a mapping between the latent representation in the encoder, and the camera poses.
A key limitation of these approaches is that they indiscriminately map latent representations to poses.
More explicitly, the network may learn to map a parked red corvette to a set of camera poses regardless of the fact the car may move.
As such, the network may not be able to successfully recover the poses when the car moves or indeed predict them at different locations where red corvettes exist.
Work by Barnes \etal \cite{barnes2018driven} has attempted to explicitly remove objects using an ephimerality mask, but relies on pre-mapped areas using LiDAR. 
While invariance to such salient objects might be achieved with sufficient data, acquiring it is not always feasible.
In this work, we propose to explicitly remove such transient objects from the latent space of a network before it is ever trained for localisation.
Such a network would have the ability to learn to regress poses in areas with large amounts of distractors, such as car parks and crowded areas, with relatively small amounts of data.

\vspace{-0.05cm}
\section{Methodology}\label{sec:meth}

It is tempting to think of neural blindness as a two-step process of \textit{identification} and \textit{removal}.
However, the purpose of this paper is to achieve a subtle and more complicated task. 
Class-specific neural blindness requires the neural network to be fundamentally incapable of representing those classes in its latent space.
Therefore, identifying the object before it can be removed fundamentally defeats the purpose, as the object has to be present in the latent space for removal to occur in the generated image.
Our architecture, loss functions and pre-training strategies force the network to learn how to represent  images while removing the presence of specific classes from its latent space. 

\subsection{Architecture}\label{sec:arch}

\begin{figure}
  \centering
  \includegraphics[width=1.0\linewidth]{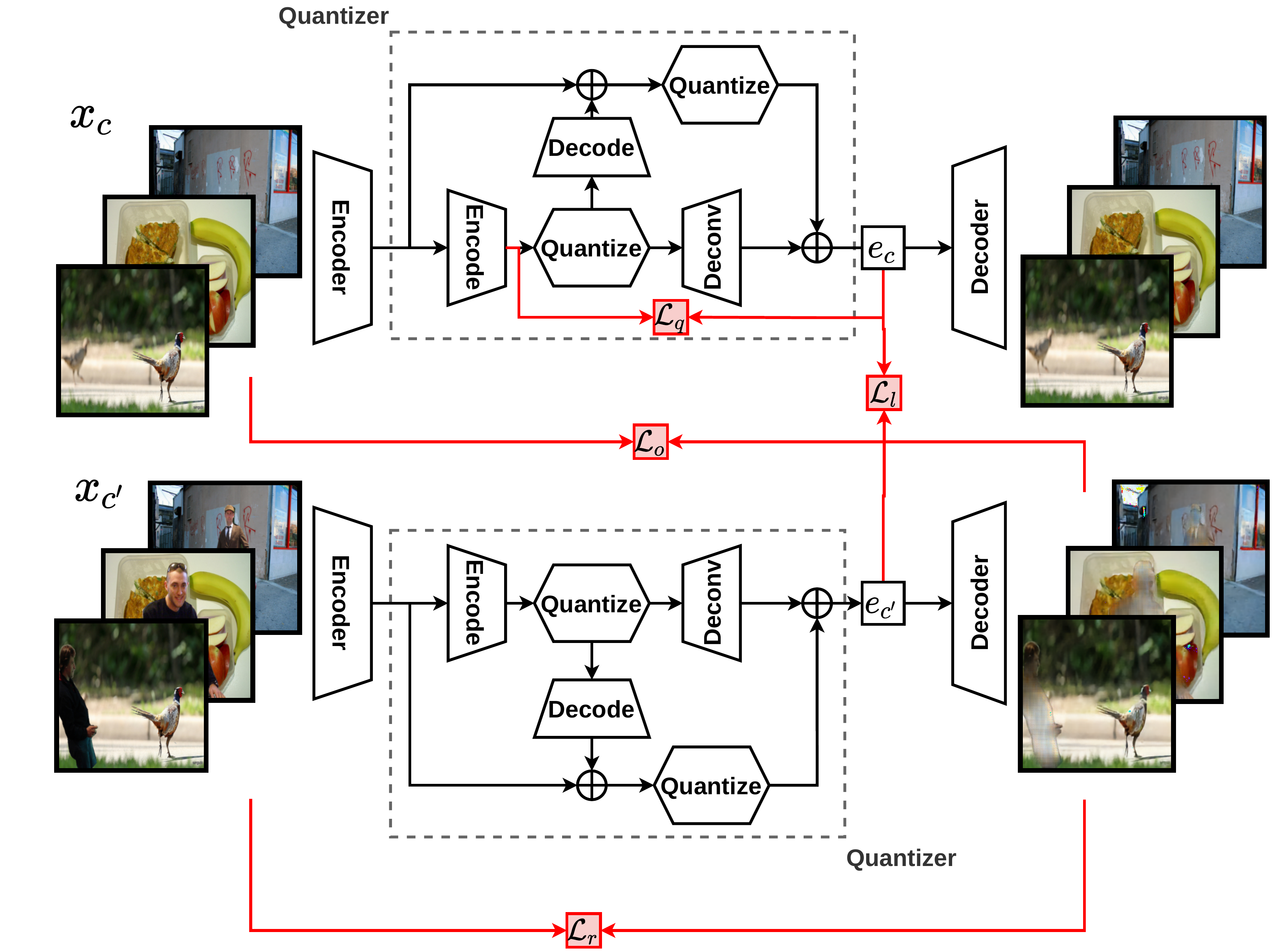} 
  \caption{System Architecture, Siamese Hierarchical \ac{vqvae}. We use images (\ensuremath{x_c}) overlaid with classes (\ensuremath{c}) to generate images \ensuremath{x_{c'}} and force the network to learn a representation (\ensuremath{e_{c'}}) that is blind to \ensuremath{c}. }
  \label{fig:sys}
  \vspace{-0.25cm}
\end{figure}

Our goal is to be able to remove a chosen class \ensuremath{c \in \mathcal{C}}, where \ensuremath{\mathcal{C}} is the set of all classes in the dataset, from the latent space of the network.
In order to accomplish this, we adopt a Siamese architecture where each arm is a \ac{vqvae}, shown in Figure \ref{fig:sys}.
This means that we can feed the network two versions of every image: the original image \ensuremath{x_c} that does \textit{not} contain \ensuremath{c} and \ensuremath{x_{c'}} where the \ensuremath{c} has been added to the image.
Since \acp{vqvae} are capable of producing both an informative embedding and a crisp reconstruction, we can compare the two arms of the network to ensure that the arm that processes \ensuremath{x_{c'}} does not learn to represent \ensuremath{c}.
We use a \ac{vqvae} as we have found, empirically,  that a standard \ac{vae} struggles produce the informative embedding and crisp reconstruction required for training.

The \acp{vqvae} employed in the Siamese architecture are two-level hierarchical, which means they each sample from two trained code-books. 
The quantised output of both code-books are then concatenated into a single latent space, shown in Figure \ref{fig:sys} as \ensuremath{e_c} and \ensuremath{e_{c'}} for \ensuremath{x_c} and \ensuremath{x_{c'}}, respectively.
These subsets of the code-book are fed to the decoder arms in order to generate the output images.
Employing a hierarchical model allows our approach to specialise the code-book such that the top code-book encodes low-frequency image detail while the bottom encodes high-frequency detail.
Since the code-books are trained to focus on different image ``levels'' it is easier to discriminate for class-specific elements.

In order to ensure the code-book is trained to correctly encode the images, we use a quantisation loss to train our latent space.
Given an input image \ensuremath{x_c}, encoder \ensuremath{E}, quantiser \ensuremath{Q} and decoder \ensuremath{D}, the loss function for the code-book is 
\begin{equation}\label{eq:lq}
 \mathcal{L}_{q}\left(x_c\right) = \left|\left| s_g[E(x_c)] - e_c ||_{2} + \beta||s_g[e_c]-E(x_c)\right|\right|_{2}
\end{equation}
where \ensuremath{e_c = Q(E(x))} is the quantised encoder output and \ensuremath{s_g} defines a stop-gradient operation that prevents back-propagation.
The first term in equation \ref{eq:lq} ensures the code-book learns to represent the input correctly, while the second term ensures the code-book does not keep changing and instead ``commits'' to a set of vectors.
Therefore, \ensuremath{\beta} describes how reluctant the network is to update its code-book.
We also adopt the standard practice of \cite{Vandenoord2018vqvae} and replace the first term of the equation with an exponential moving average update step for the code-book. 

A standard implementation of a \ac{vqvae} would combine this loss with a reconstruction loss defined as
\begin{equation}\label{re:rs}
 \mathcal{L}_{r}(x_c, e_c) = ||x_c - D(e_c)||_{2}
\end{equation}
which forces the network to learn a code-book that accurately reconstructs the image.
However, our objective is to ensure the network does \textit{not} encode a certain class \ensuremath{c} into its code-book.
Therefore, we modify the loss in equation \ref{re:rs} as:
\begin{equation}\label{re:rm}
 \mathcal{L}_{r}(x_{c'}, e_{c'}) = || (x_{c'} - D(e_{c'})) \odot (1 - \mask{x}{c'} ) ||_{2}
\end{equation}
where \ensuremath{\odot} denotes element-wise multiplication and \ensuremath{\mask{x}{c'}} is a mask for the class \ensuremath{c} we want to ignore.
More explicitly, we ensure the network gets no reward for reconstructing the target class in the image.

The final \ac{vqvae} loss can then be defined as
\begin{equation}
 \mathcal{L}_{\mysmallmath{VQ}} = \mathcal{L}_{r} + \gamma_q \mathcal{L}_{q} 
\end{equation}
where \ensuremath{\gamma} is a hyperparameter.

These losses ensure that the network is capable of learning a code-book that can successfully reconstruct an image.
However, while there is no reward for reconstructing the class \ensuremath{c}, there is nothing in these loss functions that actively prevents the network from learning global features that can reconstruct the target class.
This can happen because the network is never actively discouraged from learning the object representation.
Therefore, a loss is required to actively prevent the network from learning to encode and reconstruct the target classes.
To accomplish this, we use a Siamese loss consisting of two individual losses.
The first, ensures that the objects are not represented in the latent space, the second, that the objects are not present in the reconstructed image.

We assume the existence of an image pair \ensuremath{x_c} and \ensuremath{x_{c'}} such that the only difference between the two images is the presence of the target classes (\ensuremath{c}), as shown in Figure \ref{fig:sys}.
This is a strong assumption, and we will provide more details of the creation of such a dataset in section \ref{sec:res_data}.

Given \ensuremath{x_c} and \ensuremath{x_{c'}}, each image can be passed through one of the Siamese encoder networks such that \ensuremath{e_c = Q(E(x_c))} and \ensuremath{e_{c'} = Q(E(x_{c'}))}.
The encoded latent spaces then define a loss function as
\begin{equation}\label{re:sml}
 \mathcal{L}_{l}(e_c, e_{c'}) = \left|\left| e_{c'} - e_c \right|\right|_{1}
\end{equation}
which maintains a level of separation between the dimensions of the latent space. 

This loss ensures that the network learns to map the images containing the target classes \ensuremath{c} into a representation that fundamentally does not encode it.
One of the key insights in our work is that this type of Siamese loss is particularly well suited to a \ac{vqvae}'s architecture.
In a standard \ac{vae}, the encoder would be forced to implicitly learn to map an image into a latent space without the target classes.
In our Siamese \ac{vqvae}, the network learns to sample from the codebook such that \ensuremath{c} does not end up in the representation space.

As a final step, we use the fact that both the original image \ensuremath{x_{c}} and the reconstructed image \ensuremath{D(Q(E(x_{c'})))} should \textit{not} contain the target classes.
Therefore, we can estimate a Siamese reconstruction loss as
\begin{equation}\label{re:ro}
 \mathcal{L}_{o}(x_{c}, e_{c'}) = \left|\left| \mypar{x_{c} - D(e_{c'})} \odot \mask{x}{c'}  \right|\right|_{2}
\end{equation}
which encourages the network not to reconstruct the chosen class and is complimentary to \ensuremath{\mathcal{L}_{r}}.

The final Siamese loss can therefore be defined as
\begin{equation}
 \mathcal{L}_{S} = \mathcal{L}_l + \gamma_o \mathcal{L}_{o} 
\end{equation}
where \ensuremath{\gamma_o} is a hyperparameter defining the weight of the reconstruction loss.

Finally the losses are combined
\begin{equation}
 \mathcal{L} = \mathcal{L}_{\mysmallmath{VQ}} + \omega \mathcal{L}_{\mysmallmath{S}} %+ \omega_2 \mathcal{L}_{A}
\end{equation}
where \ensuremath{\omega} are training hyperparameters.
The combined loss function \ensuremath{\mathcal{L}} is capable of training a neural-blind architecture end-to-end, as will be shown in the following section.

\subsection{Latent Space Pre-Training} \label{sec:res_data}
As mentioned in section \ref{sec:arch}, we assume the existence of an image pair \ensuremath{x_c} and \ensuremath{x_{c'}} where the presence of the target class (\ensuremath{c}) is the only difference. 
In principle, this would limit our approach to datasets where the target classes are added and removed from scenes while the images are being captured. 
Instead, we propose to generate this from pre-existing datasets in order to train and evaluate our approach.
Given a set of object classes \ensuremath{c \in \mathcal{C}}, where \ensuremath{\mathcal{C}} is the set of all classes in the dataset (\ensuremath{\imgset{}}), we split the dataset into a subset of images that contain the category \ensuremath{\imgwth \in \imgsetwth} and images that do not \ensuremath{\imgwto \in \imgsetwto} such that \ensuremath{\imgset{c} \cap \imgset{c'} = \emptyset} and \ensuremath{\imgset{c} \cup \imgset{c'} = \imgset{}}.
For every image \ensuremath{\imgwto} that does not contain the chosen class, we randomly select an image \ensuremath{\imgwth} that does and use a mask \ensuremath{\mask{z}{c'}} to overlay one of the class objects into \ensuremath{\imgwto}. This allows us to define an overlaid image as
\begin{equation}
 \imgovr = O(\imgwto, \imgwth, \mask{z}{c'})
\end{equation}
where \ensuremath{O(\cdot)} is the function that overlays a category from \ensuremath{\imgwth} to \ensuremath{\imgwto}.
Using this overlay, a set of 3 images can be created: the original image \ensuremath{\imgnovr = \imgwto}, the original image with a category overlaid \ensuremath{\imgovr = O(\imgwto,\imgwth, \mask{z}{c'})} and a mask image \ensuremath{\mask{x}{c'}} which defines the location of the overlaid class on \ensuremath{\imgovr}. These overlay images can be used in equations \ref{re:sml} and \ref{re:ro}.
At training time, these images are automatically generated therefore randomizing the overlay at every iteration.
Although the images are sufficient to train the system to be blind, the network is never exposed to the classes in their natural context which may limit the networks ability to generalise outside of the training data.
To address this concern, we show the network natural images while training, which can be used in equation \ref{re:rm} to help the network generalise to natural data.

As previously mentioned, during training time, this dataset is dynamic, each iteration presents a new opportunity for the classes to be overlaid on different images, preventing the network from overfitting to the overlay while ensuring performance on natural images.

\section{Results}\label{sec:res}
In this section, we describe the use of a pre-trained latent space for an autonomous vehicle localisation task where blindness to visual clutter (cars) improves localisation performance. 
We also demonstrate qualitative ``blindness'' to objects by decoding latent spaces where the object has been removed.
\vspace{-0.1cm}
\subsection{Localisation Task} \label{sec:res_task}

\begin{figure}
  \centering
  \includegraphics[width=1.0\linewidth]{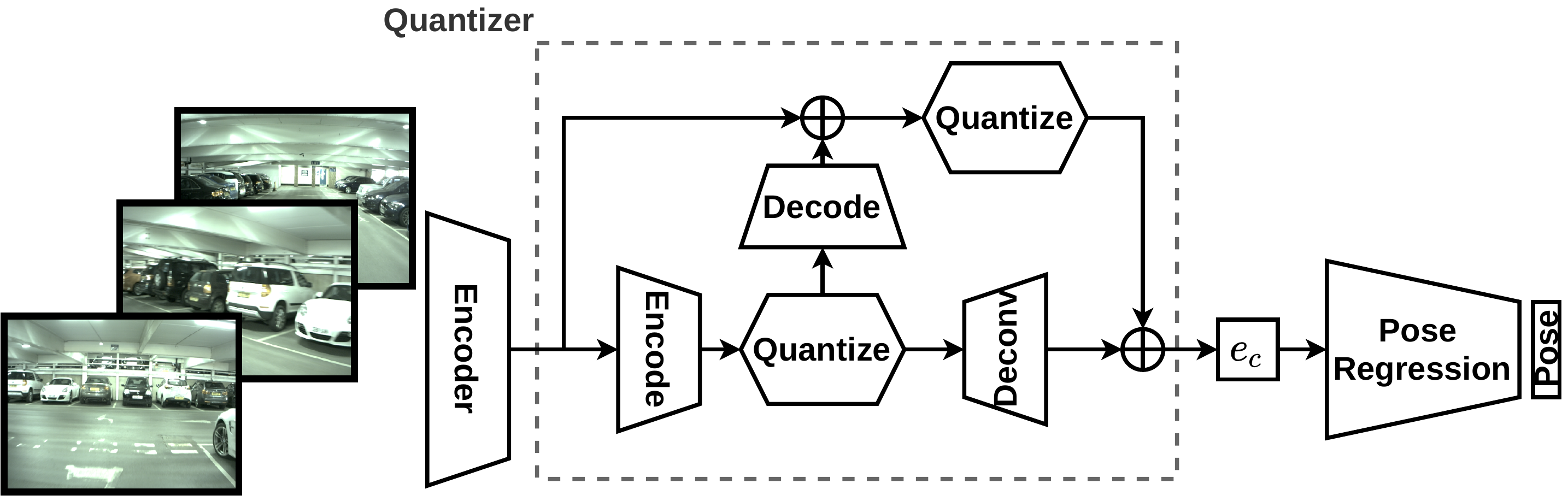} 
  \caption{VQ-VAE PoseNet Architecture, we use our Neural Blind representation \ensuremath{e_{c'}} as input to a pose-regression network to regress poses that are invariant to distractors.}
  \label{fig:vqpnet}
  \vspace{-0.5cm}
\end{figure}

A useful downstream task for neural blindness is localisation. 
Most learning-based localisation techniques employ an encoder to map an image to a latent space, then regress the pose of the camera. 
However, such approaches to localisation can easily overfit and often fail to generalise to new sequences. 
For example, training a system to learn pose from sequence A and then regress the correct pose on unseen sequence B means the latent space must be capable of overcoming environmental or structural changes between the two sequences.
Consider elements of the scene that easily change such as people in crowded scenes or cars in automotive datasets. 
These factors need to be ignored so that localisation can focus on the consistent structure of the environment.

Pose-regression networks should be able to learn to ignore distractors.
However, it isn't always feasible to capture sufficient data representing the variability needed.
In this section, we demonstrate that our neural-blind network is capable of mapping images to a more useful feature space by actively ignoring distractors, allowing us to train a distractor-robust pose-regression network \textit{without providing variance in the training data} and potentially allowing one-shot training.

\subsubsection{Car-Park Localisation}
\begin{figure}
  \centering
  \begin{tabular}{c@{}c@{}c@{}c}

  \oscarcentered{D1T1} &
  \oscarcentered{D1T2} &
  \oscarcentered{D2T1} \\
      
  \oscarcentered{\includegraphics[width=0.25\linewidth]{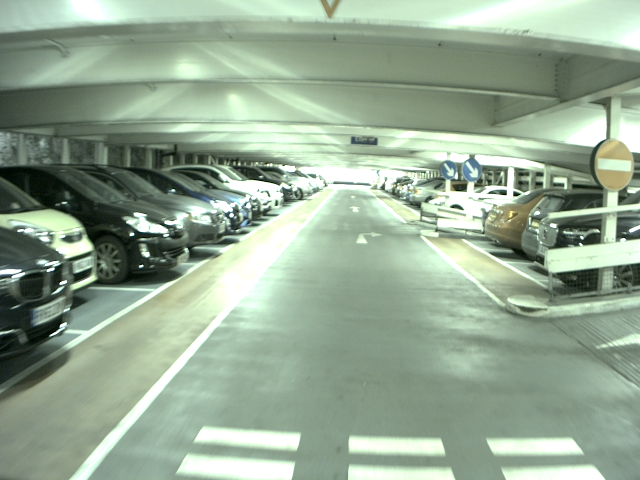}} & \hspace{-4mm}
  \oscarcentered{\includegraphics[width=0.25\linewidth]{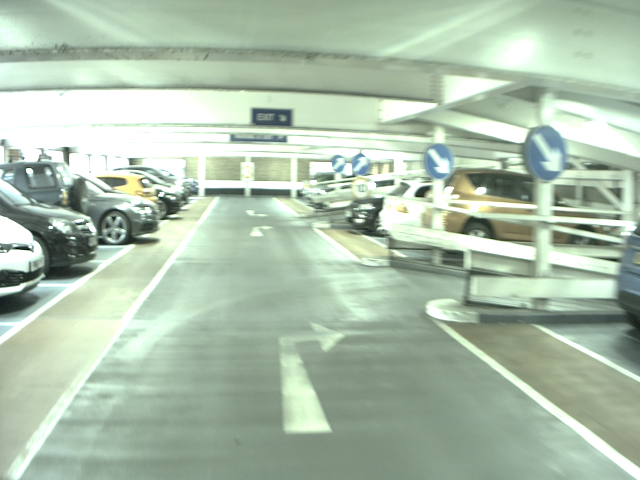}} & \hspace{-4mm}
  \oscarcentered{\includegraphics[width=0.25\linewidth]{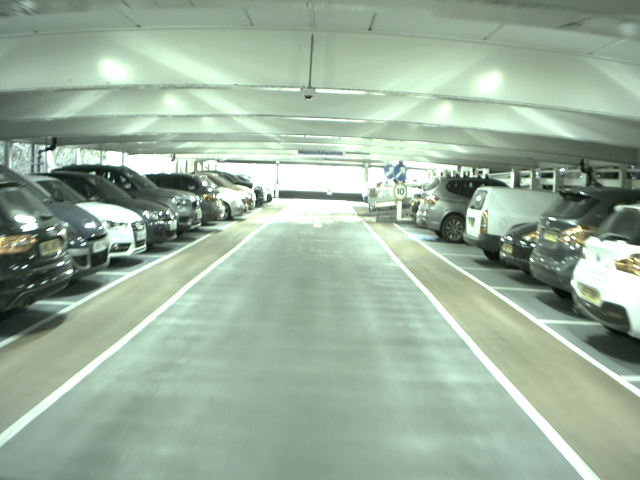}} \\
  
  \oscarcentered{\includegraphics[width=0.25\linewidth]{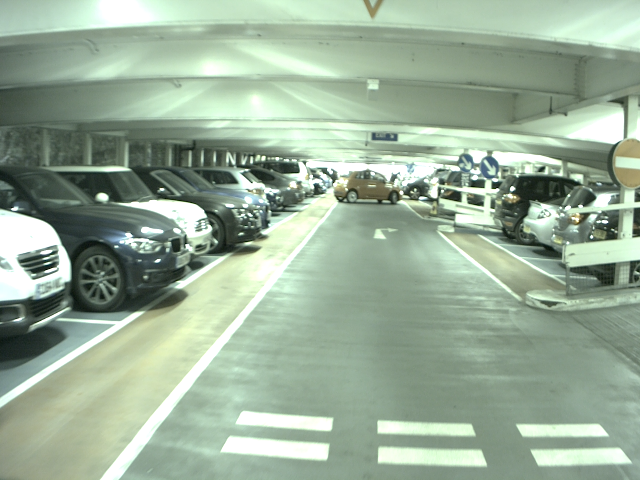}} & \hspace{-4mm}
  \oscarcentered{\includegraphics[width=0.25\linewidth]{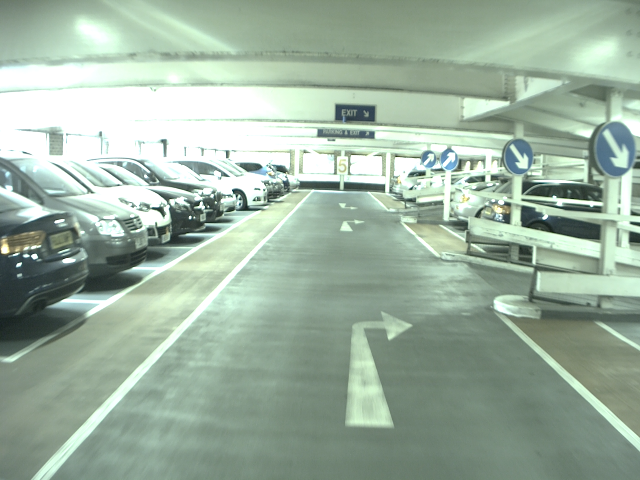}} &  \hspace{-4mm}
  \oscarcentered{\includegraphics[width=0.25\linewidth]{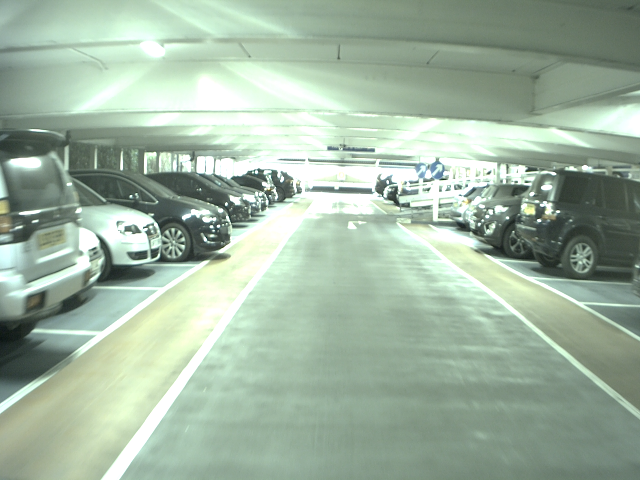}} \\ 

  \end{tabular}
  \caption{Sample Dataset images, each row is a different floor. Note the cross-floor similarity only broken by the vehicles in the scene.}
  \label{fig:avp_data}
\end{figure}
We first focus on localisation of a car in a multistory carpark. 
This is an ideal use-case for pose-regression as indoor carparks are often GPS denied. 
Furthermore, the very nature of a car park means that much of the visual input to any localisation system will include other cars, all of which have different appearance and change on a daily basis. 
Therefore, any localisation system needs to ignore cars and base its localisation on the structure of the building (as humans do). 

We first pre-train a neural-blind network that cannot see cars using the MSCOCO dataset.
Once the network is pre-trained, we fix the encoder and feed the representation space to a pose-regression network in order to estimate 6-\ac{dof} pose estimates for each input image.
This architecture can be seen in Figure \ref{fig:vqpnet}.
It should be noted that the encoder part of the network is not fine-tuned during training of the pose regressor, nor are masks of the distractors required. Instead, we simply learn to regress a pose from a vehicle-blind representation space.
Regressing poses from a representation that is incapable of encoding vehicles should lead to better one-shot learning and pose regression. 
\begin{table}[]
\centering
\resizebox{\linewidth}{!}{%
\begin{tabular}{|l||c||c|c|c|}
 \hline
    & \multicolumn{1}{l||}{\textbf{Training Sequence}}  & \multicolumn{1}{l|}{\textbf{D1T1}} & \multicolumn{1}{l|}{\textbf{D1T2}} & \multicolumn{1}{l|}{\textbf{D2T1}} \\ \hline
PoseNet \cite{kendall2017geometric}                                           &  \underline{0.59}    &   8.07             &   8.61             &   12.99            \\ \hline
\begin{tabular}[c]{@{}l@{}}Non-Blind VQ-VAE \\ PoseNet (Us)\end{tabular}      &   1.15             &   8.68             &   8.66             &   13.74            \\  \hline
\begin{tabular}[c]{@{}l@{}}Car-Blind VQ-VAE \\ PoseNet (Us)\end{tabular}      &   1.08             &   \textbf{6.87}    &   \textbf{7.42}    &   \textbf{12.46}   \\  \hline
\end{tabular}%
}
\caption{Comparison of Median Error, in meters, between PoseNet\cite{kendall2017geometric}, VQ-VAE based PoseNet and a VQ-VAE PoseNet trained to be blind to cars.\vspace{-0.5cm}}
\label{tab:floors}
\end{table} 
To train the pose regression network we use a dataset that consists of three separate runs through a 6-floor multi-storey carpark. The runs were captured over two days, with two captures on the first day and a third capture on the second day. 
Each of these sequences traverses the entire 6 floors. We will refer to these sequences as D1T1, D1T2 and D2T1 to represent the Day (D) and Trajectory (T).
The capture vehicle was equipped with 3 front-facing cameras and a 16-beam \acs{lidar}.
Of the three front facing cameras, there two are narrow field-of-view cameras (Left and Right) and one wide-angle (Centre) which we use as input to the pose regressor.
The vehicle traversed 6 floors, over \ensuremath{6,500 m} of driving and a volume of \ensuremath{98500 m^3}.
We use the \acs{lidar} to create the ground truth poses and extract between 2500 and 4000 images per-camera, per-trajectory. 

\subsubsection{Cross-Floor Validation}
In this experiment we exploit the fact that the floors of the multistory carpark are structurally identical, as can be seen in Figure \ref{fig:avp_data}.
However, despite structural similarity of the carpark, the appearance varies drastically due to the different population of parked cars. 
To test the effect of neural blindness we trained three different networks (Blind, Non-Blind and PoseNet \cite{kendall2017geometric}) on the second floor of the multistory carpark and evaluated 2D localisation performance for floor one across D1T1, D1T2 and D2T1. 

We train a neural-blind pose regression network for 100 epochs, with a learning rate of \ensuremath{0.0001} using an ADAM optimizer and a step learning rate scheduler with \ensuremath{\gamma = 0.5}. 
We use the same hyperparameters to train a non-blind network. 
As in other work, \cite{Kendall2015Posenet}, we report the median error in meters and evaluate our networks against a state-of-the-art PoseNet \cite{kendall2017geometric} trained on the same data. 
All networks are trained using a homescedastic loss (\cite{kendall2017geometric}). 
For PoseNet, we use a pre-trained ResNet as an encoder and allow the network to fine-tune the entire architecture. 
For our networks, we do not fine-tune the encoders, and expect the network's representation space to be robust enough for pose regression. 

As can be seen in Table~\ref{tab:floors}, PoseNet has the lowest training error suggesting it can memorise the images. 
However, while the Non-blind \ac{vqvae} has comparable performance to PoseNet on the unseen sequences (D1T1, D1T2, D2T1), the Car-blind \ac{vqvae} has a significantly reduced localisation error. 
This demonstrates the benefit of a localisation framework that is blind to the distractors (cars) in the environment. 

In a second experiment, we train both a neural blind \ac{vqvae} Pose regression network and a standard PoseNet on the entire D1T2 (6 Floors) sequence to regress the full 6 degree of freedom vehicle pose. 
We train both networks for 500 epochs. 
We then test both networks on the full 6-floors, using unseen sequences D1T1 and D2T2.
Tables \ref{tab:traj1} and \ref{tab:traj2} show the performance of PoseNet \cite{kendall2017geometric} as well as our approach.
Our Neural-Blind Pose regression can easily outperform the standard PoseNet, reducing the error by a significant margin.
This is because PoseNet cannot identify the cars as distractors. 
In this scenario, PoseNet learns to map car-features to poses, and when those features are modified (or absent), the localisation performance suffers.
In contrast, our approach doesn't see or  encode car-features.
Therefore, the network must rely on other features to estimate the pose and does not suffer in the presence of high variability of distractors.
Since any class can be removed from the latent space, we expect this approach to work in any scene where visual clutter is a problem.
\begin{table}
\centering
\resizebox{\linewidth}{!}{%
\begin{tabular}{|l|c|c|c|}
\hline
                                                     & \multicolumn{1}{|l|}{Left} & \multicolumn{1}{|l|}{Right} & \multicolumn{1}{|l|}{Centre-Wide} \\ \hline
PoseNet \cite{kendall2017geometric} & 1.64                     & 1.74                      & 1.74            \\ \hline
Neural-Blind Pose (Us)                            &  \textbf{0.99}                     &  \textbf{1.08}                      &  \textbf{1.09} \\ \hline
\end{tabular}%
}
\caption{Median Error on D1T1.}
\label{tab:traj1}
\vspace{-0.25cm}
\end{table}
\begin{table}
\centering
\resizebox{\linewidth}{!}{%
\begin{tabular}{|l|c|c|c|}
 \hline
                                                     & \multicolumn{1}{|l|}{Left} & \multicolumn{1}{|l|}{Right} & \multicolumn{1}{|l|}{Centre-Wide} \\ \hline
PoseNet \cite{kendall2017geometric} & 2.16                     & 2.34                      & 2.24                            \\ \hline
Neural-Blind Pose (Us)                            &  \textbf{1.61}                     &  \textbf{1.74}                      &  \textbf{1.62} \\ \hline                           
\end{tabular}%
}
\caption{Median Error on D2T1.\vspace{-0.1cm}}
\label{tab:traj2}
\vspace{-0.5cm}
\end{table}
\vspace{-0.15cm}
\subsection{Blindness Decoding}\label{sec:res_inp}
\begin{figure}
  \centering
  \includegraphics[width=0.15\linewidth]{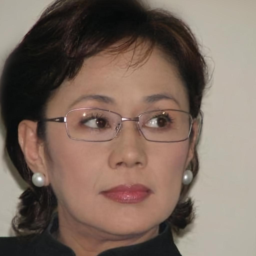}
  \includegraphics[width=0.15\linewidth]{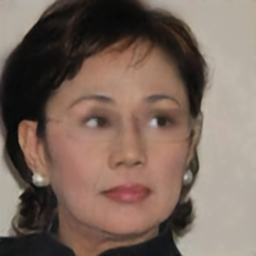}
  \includegraphics[width=0.15\linewidth]{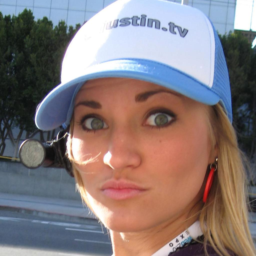}
  \includegraphics[width=0.15\linewidth]{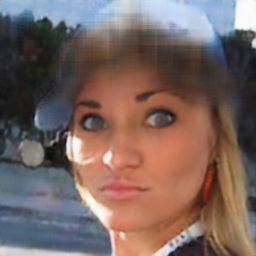}
  
  \includegraphics[width=0.15\linewidth]{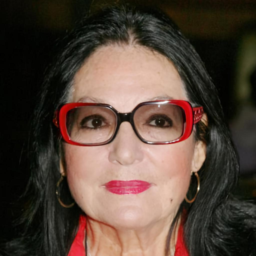}
  \includegraphics[width=0.15\linewidth]{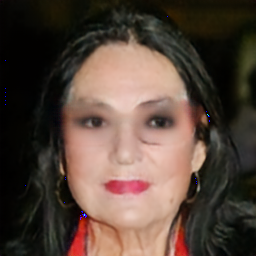}
  \includegraphics[width=0.15\linewidth]{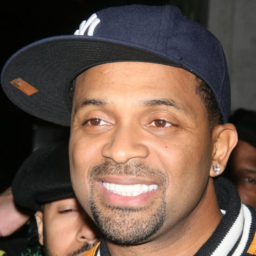}
  \includegraphics[width=0.15\linewidth]{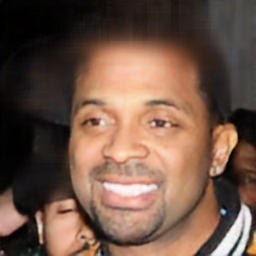}
  
  \includegraphics[width=0.15\linewidth]{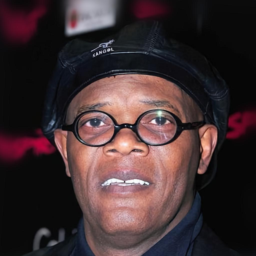}
  \includegraphics[width=0.15\linewidth]{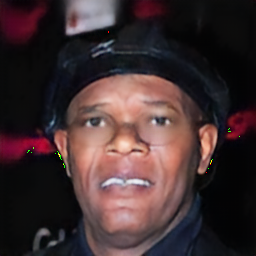}
  \includegraphics[width=0.15\linewidth]{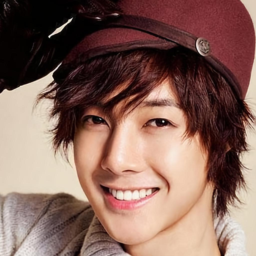}
  \includegraphics[width=0.15\linewidth]{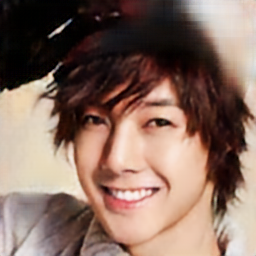} 
  \caption{Facial Modification: Removing items. Note approach has not been optimised for appearance.\vspace{-0.3cm}}
  \label{fig:glasses}
  \vspace{-0.5cm}
\end{figure}
We now demonstrate that our network becomes blind to certain classes by decoding the latent space of a neural-blind \ac{vqvae}.
We show, qualitatively, that the class the network is blind to is not reconstructed. 
It should be noted that our network is not rewarded for the quality of the image ``inpainting'' but is simply incapable of representing these classes in its latent space, the result is in-painting by proxy.

In a first example, we show how our neural-blind \ac{vqvae} can be trained to become blind to facial apparel, particularly glasses and hats.
We train two blind \acp{vqvae} (glasses and hats) using the CelebAMask-HQ dataset \cite{CelebAMask-HQ}.
While many approaches have shown excellent results in this domain \cite{liu2018b}, they all require a mask before they remove the features.
In our work, the network is fundamentally incapable of representing these categories so a mask is not necessary at runtime. 
As can be seen in Figure \ref{fig:glasses}, our network is capable of removing the eyeglasses class while maintaining the eyes and facial expression.
It is also capable of removing hats while maintaining sensible hair appearance.
It should be noted that the network was NOT provided with a mask, nor has the network been trained to optimize for appearance (using an adversarial loss or similar).
Instead, the network is incapable of encoding these classes into its latent representation.

\begin{figure}
  \centering
  \captionsetup[subfigure]{justification=centering,singlelinecheck=false,font=small}
  
    \begin{tabular}{rl}
        \oscarcentered{\small Car}&\hspace{-6mm}
        \begin{tabular}{ccc}
        \oscarcentered{\includegraphics[width=0.20\linewidth]{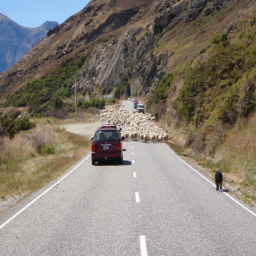}} & \hspace{-8mm}
        \oscarcentered{\includegraphics[width=0.20\linewidth]{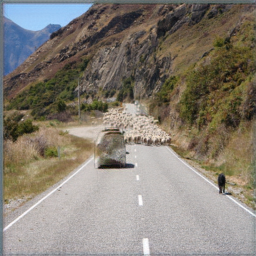}} & \hspace{-8mm}
        \oscarcentered{\includegraphics[width=0.20\linewidth]{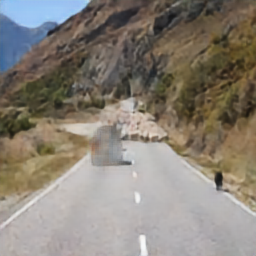}} 
        \end{tabular}\\

        %Person
        \oscarcentered{\small Person} &\hspace{-6mm}
        \begin{tabular}{ccc}
        \oscarcentered{\includegraphics[width=0.20\linewidth]{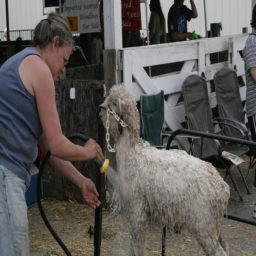}} & \hspace{-8mm}
        \oscarcentered{\includegraphics[width=0.20\linewidth]{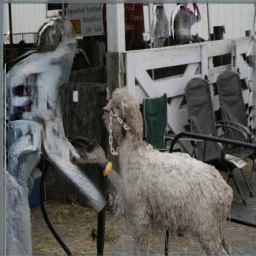}} & \hspace{-8mm}
        \oscarcentered{\includegraphics[width=0.20\linewidth]{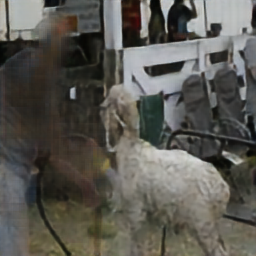}} 
        \end{tabular}\\

        %Fire Hydrant
        \oscarcentered{\small Fire \\ Hydrant} &\hspace{-6mm}
        \begin{tabular}{ccc}
            \oscarcentered{\includegraphics[width=0.20\linewidth]{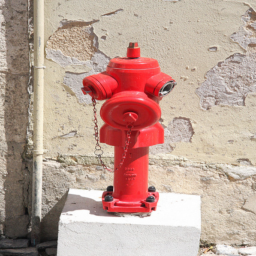}} & \hspace{-8mm}
            \oscarcentered{\includegraphics[width=0.20\linewidth]{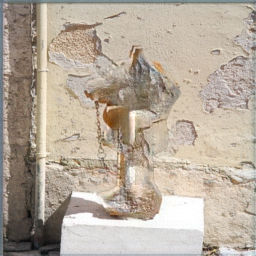}} & \hspace{-8mm}
            \oscarcentered{\includegraphics[width=0.20\linewidth]{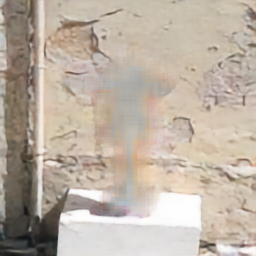}} 
        \end{tabular}\\
        
        %Cow
        \oscarcentered{\small Cow} &\hspace{-6mm}
        \begin{tabular}{ccc}
            \oscarcentered{\includegraphics[width=0.20\linewidth]{./images/natural_new/00238_nat.png}} & \hspace{-8mm}
            \oscarcentered{\includegraphics[width=0.20\linewidth]{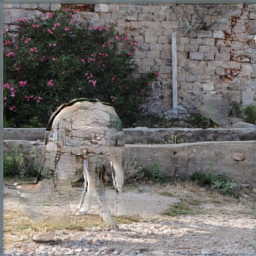}} & \hspace{-8mm}
            \oscarcentered{\includegraphics[width=0.20\linewidth]{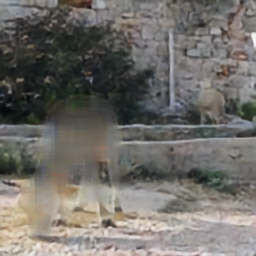}} 
            \vspace{-3mm}
        \end{tabular}\\
        %Fork
        \oscarcentered{\small Fork} &\hspace{-6mm}
        \begin{tabular}{ccc}
            \subfloat[Original Image]                                       {\oscarcentered{\includegraphics[width=0.20\linewidth]{./images/natural_new/00048_nat.png}}} & \hspace{-8mm}
            \subfloat[DeepFill V2 \cite{yu2018generative}\cite{yu2019free}  \textit{(Mask)}]     {\oscarcentered{\includegraphics[width=0.20\linewidth]{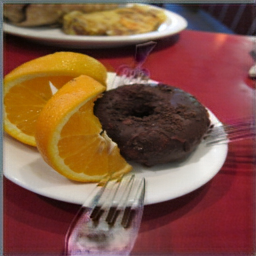}}} & \hspace{-8mm}
            \subfloat[\hspace{1mm}Ours \newline \textit{ (\textbf{No Mask})}]        {\oscarcentered{\includegraphics[width=0.20\linewidth]{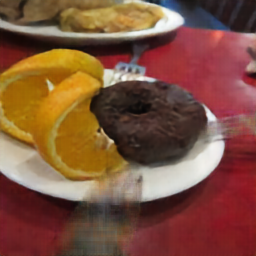}}} 
        \end{tabular}\\
    \end{tabular} 
    
  \caption{Inpainting in Natural Images: Original image, DeepFill V2 \cite{yu2018generative}\cite{yu2019free} and Ours.}
  \label{fig:mscoco_blind_nat}
  \vspace{-0.2cm}
\end{figure}

In a second example, we select a representative set of categories to be blind to \ensuremath{c=\{car, person, fire{\text{-}}hydrant, cow, fork\}} and train on each class independently using the MSCOCO 2017 dataset.
We evaluate our approach against a state-of-the-art inpainting approach (DeepFill V2 \cite{yu2018generative}\cite{yu2019free} ) and show that our approach is capable of out-performing inpainting without the use of a mask.
Figure \ref{fig:mscoco_blind_nat} shows qualitative results on natural images from the validation set of MSCOCO 2014.
The first column are the original images, the second column is DeepFill V2 \cite{yu2018generative}\cite{yu2019free}  applied to these images with a mask and the third is our neural blind \ac{vqvae} \textbf{without a mask}.
Note that comparing against an approach that is given a mask is inherently unfair to our approach, especially in situations where the mask is unavailable at runtime.
However, it shows that the reason for performance gains in localisation are down to the removal of objects in the encoder latent space.

To evaluate quantitatively, we create a fixed unseen dataset for each class out of the MSCOCO \textit{2014} validation set with randomized overlays.
This is done to ensure that there is ground-truth data for the area covered by the target class.
Natural images, such as those shown in Figure \ref{fig:mscoco_blind_nat},  do not contain this information which would make meaningful quantitative evaluation impossible. 
Note that the chosen classes provide representative performance, but our algorithm is flexible enough to learn blindness for other classes, including combinations of classes.
Table \ref{tab:inp} shows the mean \ensuremath{L_1}, \ac{mse} and \ac{psnr} estimated between the overlaid images and the originals.
It can be seen that we outperform DeepFill V2 in all metrics.
This shows that our approach is capable of robustly inpainting a chosen class even without the use of a mask.

\begin{table}
\centering 
\resizebox{0.99\linewidth}{!}{%
\begin{tabular}{l|l|ccc}
                     & \textbf{Approach}                                             & \textbf{L1}    & \textbf{MSE}   & \textbf{PSNR}  \\ \hline
\multirow{2}{*}{Person}       & DeepFill V2 \cite{yu2018generative}\cite{yu2019free}  \textit{(Mask)} & 0.214 & 0.123 & 9.68  \\
                     & \textbf{Ours \textit{(No Mask)}}                                 & \textbf{0.129} & \textbf{0.056} & \textbf{13.86} \\ \hline
\multirow{2}{*}{Car} & DeepFill V2 \cite{yu2018generative}\cite{yu2019free}  \textit{(Mask)} & 0.229          & 0.136          & 9.30           \\
                     & \textbf{Ours \textit{(No Mask)}}                                 & \textbf{0.148} & \textbf{0.069} & \textbf{13.28} \\ \hline
\multirow{2}{*}{Fire-Hydrant} & DeepFill V2 \cite{yu2018generative}\cite{yu2019free}  \textit{(Mask)} & 0.215 & 0.125 & 9.55  \\
                     & \textbf{Ours \textit{(No Mask)}}                                 & \textbf{0.102} & \textbf{0.036} & \textbf{15.66} \\ \hline
\multirow{2}{*}{Cow} & DeepFill V2 \cite{yu2018generative}\cite{yu2019free}  \textit{(Mask)} & 0.220          & 0.129          & 9.39           \\
                     & \textbf{Ours \textit{(No Mask)}}                              & \textbf{0.119} & \textbf{0.046} & \textbf{14.33} \\ \hline
\multirow{2}{*}{Fork}         & DeepFill V2 \cite{yu2018generative}\cite{yu2019free}  \textit{(Mask)} & 0.195 & 0.107 & 10.20 \\
                     & \textbf{Ours \textit{(No Mask)}}                              & \textbf{0.082} & \textbf{0.027} & \textbf{16.89} \\ \hline
\end{tabular}%
}

\caption{Per-class quantitative results on Overlay Dataset.\vspace{-0.2cm}}
\label{tab:inp}
\vspace{-0.4cm}
\end{table} 

\vspace{-0.3cm}
\section{Conclusions}
We have trained a network to \textit{not} encode certain classes into its latent space and shown this latent space can be used for downstream tasks.
This kind of ``negative attention'' is capable of revolutionising the way we think about certain tasks.
It means that visual clutter can be removed from tasks like localisation in crowded scenes or tracking in busy environments.
Our contributions allow networks to be trained with smaller amounts of data, as we can choose to remove attention from elements that may cause the network to overfit.

{\small
\bibliographystyle{ieee_fullname}
\bibliography{extracted}
}

\end{document}